\documentclass{article}

\usepackage{neurips_2023}

\usepackage[utf8]{inputenc} 
\usepackage[T1]{fontenc}    
\usepackage{hyperref}       
\usepackage{url}            
\usepackage{booktabs}       
\usepackage{amsfonts}       
\usepackage{nicefrac}       
\usepackage{microtype}      
\usepackage{xcolor}         
\usepackage{graphicx}
\usepackage{authblk}
\newcommand*{\affaddr}[1]{#1} 
\newcommand*{\affmark}[1][*]{\textsuperscript{#1}}
\newcommand*{\email}[1]{\texttt{#1}}
\usepackage{caption}
\usepackage{subcaption}

\title{Graph Transformer Network for Flood Forecasting with Heterogeneous Covariates}

\author{%
\textbf{Jimeng Shi}\affmark[1], \textbf{Vitalii Stebliankin}\affmark[1], \textbf{Zhaonan Wang}\affmark[2], \textbf{Shaowen Wang}\affmark[2], \textbf{Giri Narasimhan}\affmark[1] \\
\affaddr{\affmark[1]Florida International University} \\
\affaddr{\affmark[2]University of Illinois Urbana-Champaign} \\
\email{\{jshi008,vsteb002,giri\}@fiu.edu, \{znwang,shaowen\}@illinois.edu} \\
}

\begin{document}

\maketitle

\begin{abstract}
Floods can be very destructive causing heavy damage to life, property, and livelihoods. Global climate change and the consequent sea-level rise have increased the occurrence of extreme weather events, resulting in elevated and frequent flood risk. Therefore, accurate and timely flood forecasting in coastal river systems is critical to facilitate good flood management. However, the computational tools currently used are either slow or inaccurate. In this paper, we propose a \underline{Flood} prediction tool using \underline{G}raph \underline{T}ransformer \underline{N}etwork (FloodGTN) for river systems. More specifically, FloodGTN learns the spatio-temporal dependencies of water levels at different monitoring stations using Graph Neural Networks (GNNs) and an LSTM. It is currently implemented to consider external covariates such as rainfall, tide, and the settings of hydraulic structures (e.g., outflows of dams, gates, pumps, etc.) along the river. We use a Transformer to learn the attention given to external covariates in computing water levels. We apply the FloodGTN tool to data from the South Florida Water Management District, which manages a coastal area prone to frequent storms and hurricanes. Experimental results show that FloodGTN outperforms the physics-based model (HEC-RAS) by achieving higher accuracy with 70\% improvement while speeding up run times by at least 500x.
\end{abstract}

\section{Introduction}
Floods can result in catastrophic consequences with considerable loss of life, huge socio-economic impact \cite{wu2021new}, property damage \cite{brody2007rising}, and environmental devastation \cite{yin2023flash}.
Floods are a threat to food and water security and to sustainable development \cite{kabir2019impacts}.
What is more alarming is that studies suggest that global climate change may lead to drastically increased flood risks, in both frequency and scale \cite{wing2022inequitable,hirabayashi2013global}.
Therefore, accurately predicting flood events is of utmost importance.

Flood forecasting is traditionally achieved by building detailed physics-based computational models that take into consideration high-resolution terrain data, groundwater data, reservoirs, man-made structures, canals, and river networks \cite{sampson2015high,zang2021improving}. 
Hydrological dynamics of the entire watershed are simulated on detailed grid-based terrain elevation maps \cite{vieux2001distributed} by solving complex partial differential equations (PDEs) \cite{paniconi2015physically, yin2023physic}.
Consequently, good simulations are accurate but computationally prohibitive, especially on large watersheds \cite{hu2022accelerating, kauffeldt2016technical}.

Machine learning (ML) and deep learning (DL) have emerged as powerful approaches for cutting-edge research in this domain \cite{zeng2023global, carvalho2019systematic, mosavi2018flood}.
Many of the previous efforts on this problem have used ML/DL tools as black boxes with little or no domain knowledge incorporated in architecting a model \cite{sajedi2018novel, gizaw2016regional, motta2021mixed, sankaranarayanan2020flood, gude2020flood, shi2023deep}, and often with little or no fine-tuning.
Some of them also usually ignore the heterogeneity of the input variables and/or the geospatial relationships between the measurement sites.
Finally, almost all of them ignore useful and reliable predictions of some covariates, such as precipitation and tide \cite{delaney2020forecast, zarei2021machine}.

\begin{figure}[h]
     \centering
     \begin{subfigure}[b]{0.48\textwidth}
        \centering
        \includegraphics[scale=0.23]{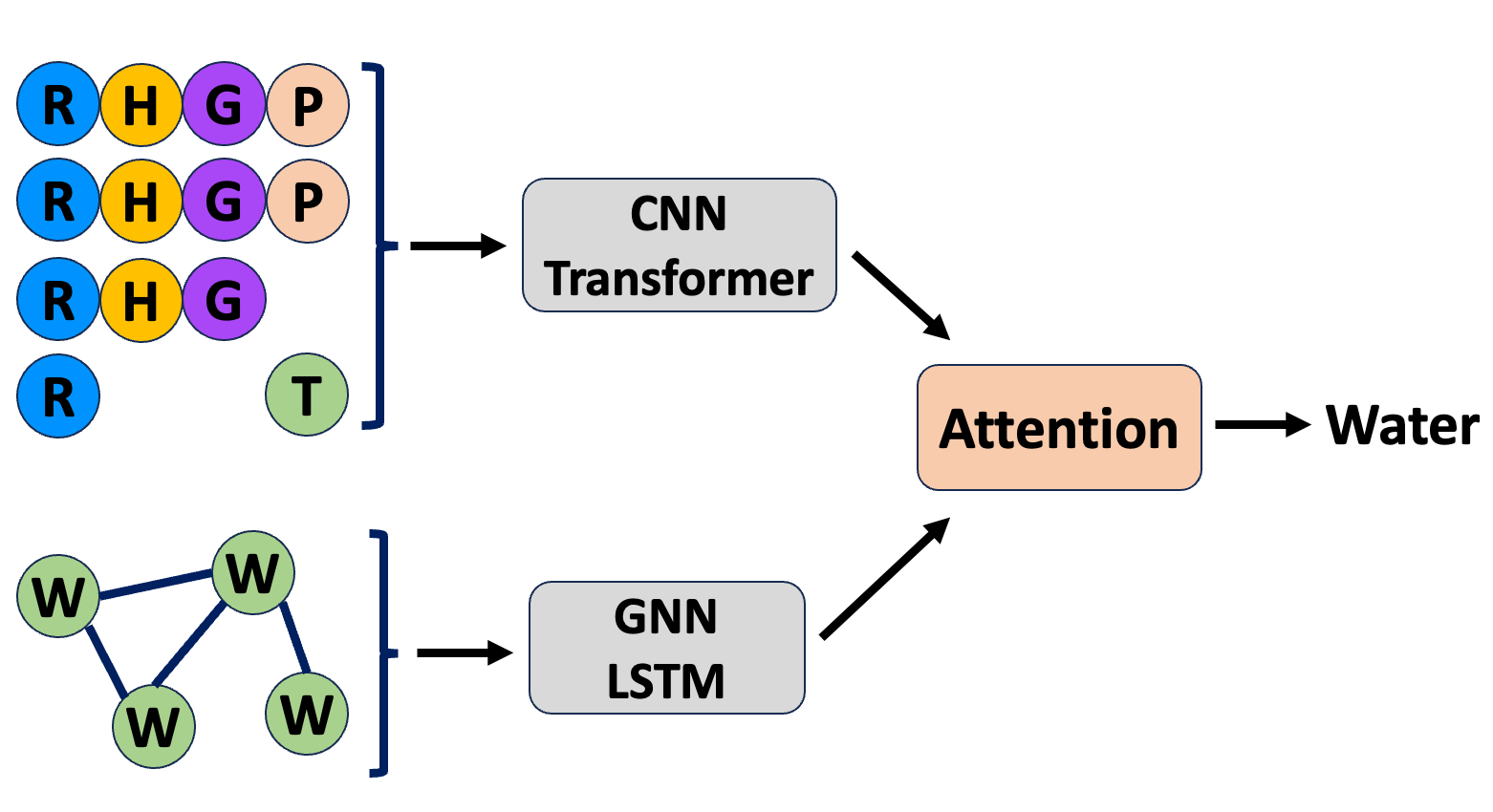}
        \caption{FloodGTN-Parallel}
        \label{fig:gtn_parallel}
     \end{subfigure}
     \hfill
     \begin{subfigure}[b]{0.51\textwidth}
        \centering
        \includegraphics[scale=0.23]{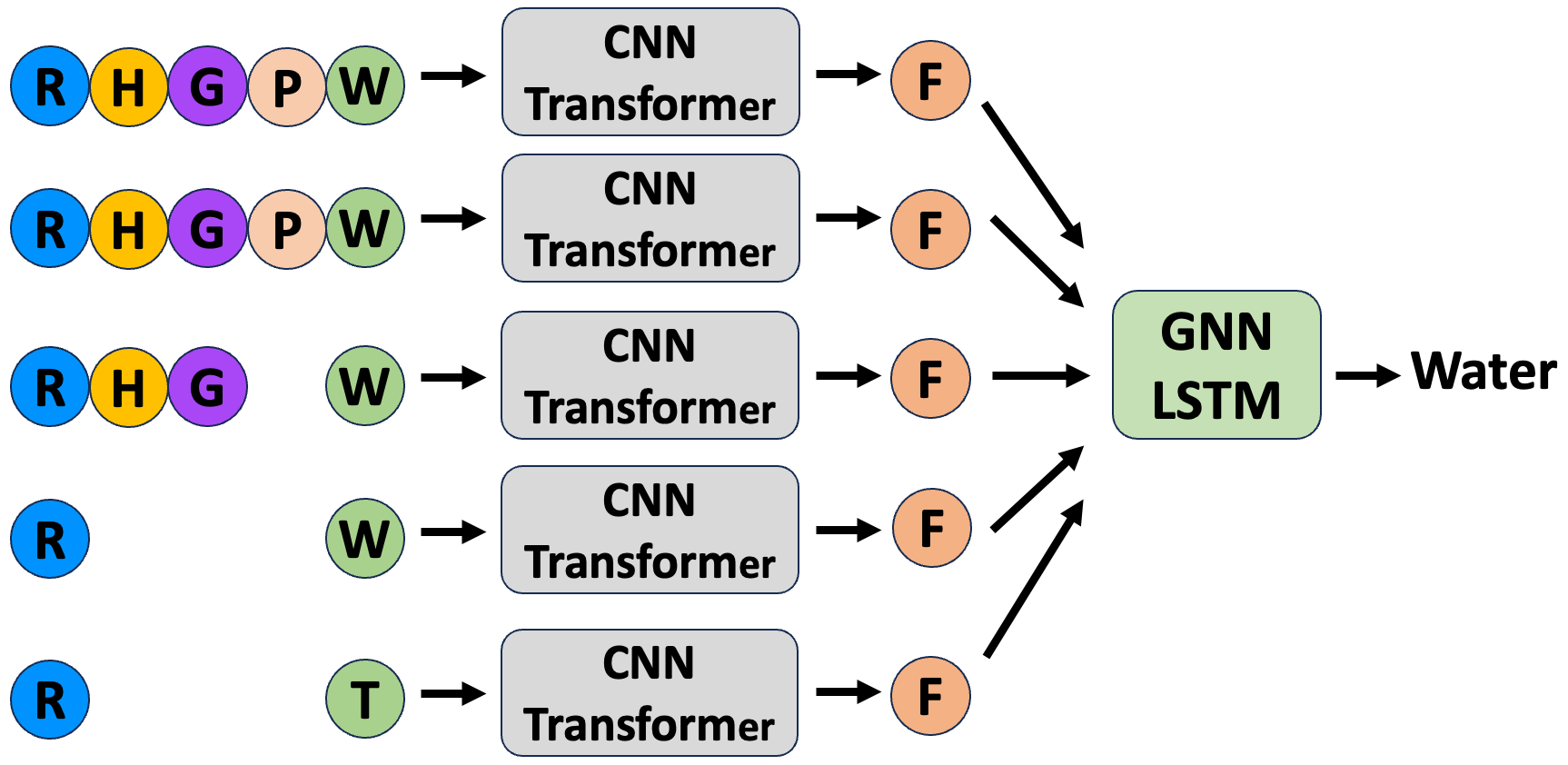}
        \caption{FloodGTN-Series}
        \label{fig:gtn_series}
     \end{subfigure}
        \caption{Two versions of FloodGTN}
        \label{fig:gtn}
\end{figure}

In this work, we propose a machine learning tool called \underline{Flood} prediction tool using \underline{G}raph \underline{T}ransformer \underline{N}etwork (FloodGTN).
We present two possible architectures for FloodGTN, one that connects the components in \emph{parallel} and one that connects it in \emph{series}, as shown in Figures \ref{fig:gtn_parallel} and \ref{fig:gtn_series}.
Details are given in Section \ref{sec:method}.
An important feature of our work is that the graph underlying the GNN component has edge connections that mirror the geospatial layout of the river system, a distinct advantage resulting from the use of GNNs, not afforded by other machine learning models.
A second important feature of our work is the use of forecast-informed covariates in our model, thus recognizing the fact that modern weather and tide predictions are reliable enough to be extremely useful in flood prediction.
Finally, the attention component of the transformers in the architecture allows us to incorporate \emph{explainability} into our models.

\section{Problem Formulation and Methodology}
\label{sec:method}
Flood forecasting will be achieved by predicting water levels, $Y$, at multiple control points in a river system for $k>0$ time steps in the future, taking inputs, $X$, from the past $w\ge 0$ time steps, while making use of additional information on external covariates that can be reliably predicted for $k$ time steps in the future. 
We train machine learning models to learn a mapping function $f$ with parameters $\theta$ from input matrix $X$ to output matrix $Y$:
\begin{equation}
    \label{eq:targets}
    f_\theta: (X_{t-w+1:t}, X^{cov}_{t+1:t+k}) \rightarrow Y_{t+1:t+k},
\end{equation}
where the subscripts of the time-series variables $X$ and $Y$ represent the time ranges under consideration, and the superscript $cov$ refers to the input covariates that are known to be reliably predicted and available for computations.
In particular, we use DL models that combine multiple approaches such as graph neural networks (GNNs), attention-based transformer networks, long short-term memory networks (LSTMs), and convolutional neural networks (CNNs) to achieve our objectives. 

\paragraph{FloodGTN-Parallel and FloodGTN-Series.}
In the parallel version, the \texttt{Transformer} learns feature representations from covariates while \texttt{GCN-LSTM} extracts spatiotemporal dynamics of water levels. \texttt{Attention} is used to figure out the contributions of each covariate to water level predictions.
The serial version is a geo-spatially aware architecture. Here the \texttt{Transformer} learns feature representations of each measurement station along the river while \texttt{GCN-LSTM} extracts information on the spatiotemporal dynamics of learned feature representations from each station.

\section{Experiments}
\begin{figure}[ht]
\centering
\includegraphics[scale=0.4]{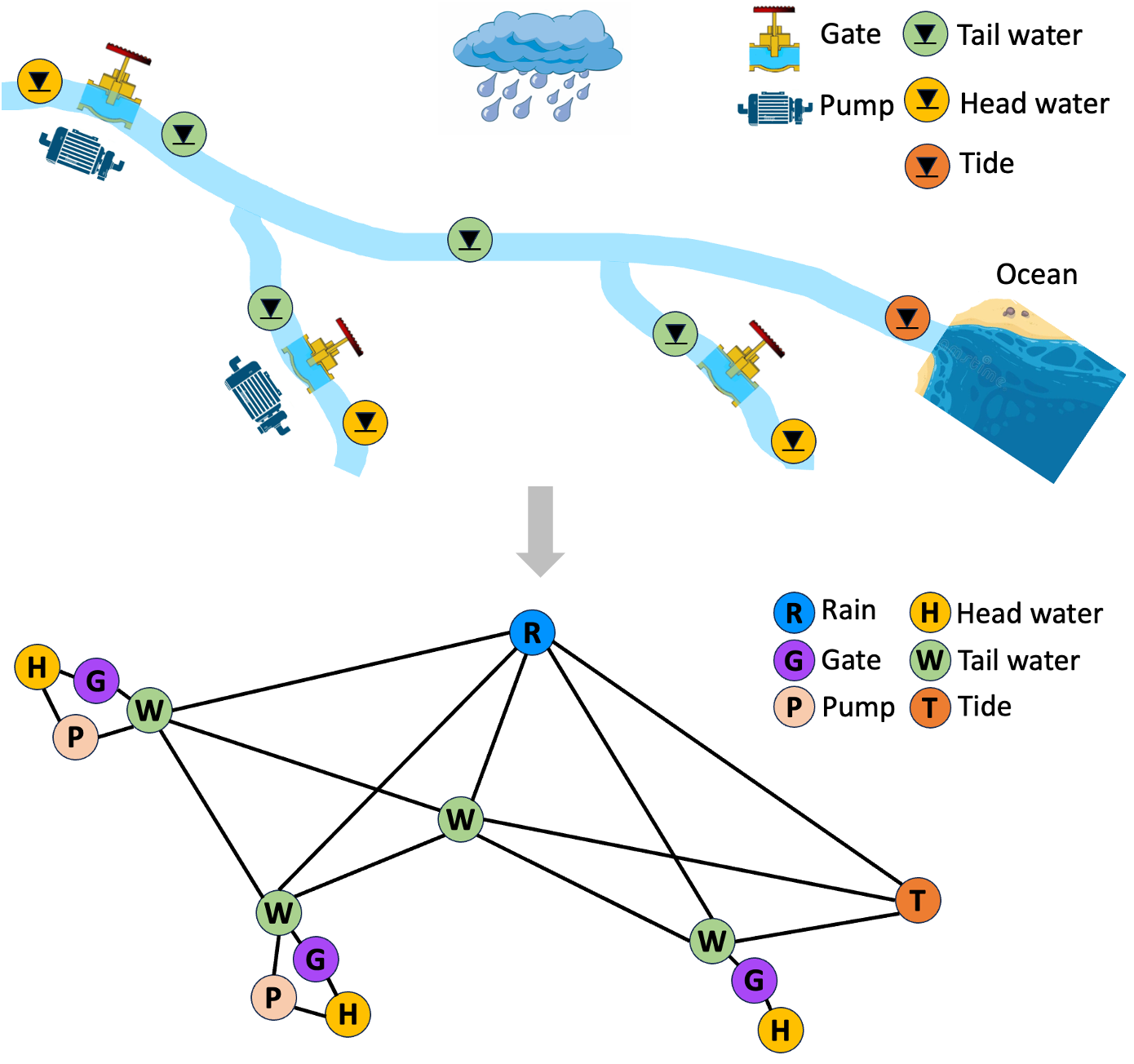}
\caption{Coastal river system in the South Florida watershed. Top: schematic diagram of river systems. Bottom: schematic diagram with symbols for better modeling representation.}
\label{fig:domain_all}
\end{figure}

\paragraph{Data.} We obtained data on the last (coastal) part of the South Florida watershed from the South Florida Water Management District's (SFWMD) DBHydro database \cite{dbhydro2023sfwmd}.
The data set we used in the work recorded the hourly observations for water levels and external covariates from 2010 to 2020.
As shown in Figure \ref{fig:domain_all}, the river system has three branches/tributaries and includes several hydraulic structures (gates, pumps) located along rivers to control water flows.
Water levels are also impacted by tidal information since the river system empties itself into the ocean.
The tool, FloodGTN, was used on this data set to predict water levels at four specific locations marked by green circles in Figure \ref{fig:domain_all}.
It is useful to note that this portion of the river system flows through the large metropolis of Miami, which has a sizable population, commercial enterprises, and an international airport in its close vicinity.

\paragraph{Experimental design.} We set $w=72$ hours and $k=24$ hours.
The sliding window \cite{li2014rolling,gidea2018topological} (also known as looking window, rolling window \cite{shi2023explainable}) strategy was used to process the entire dataset.
For deep learning models, the first 80\% and the last 20\% of the data set were used as the training and test sets, respectively.
The physics-based model (HEC-RAS) and deep learning (DL) models (other than FloodGTN) compared in this work include the following:
\begin{itemize}
  \item HEC-RAS \cite{brunner2002hec}: a physics-based model. We use a coupled 1D and 2D hydraulic model.
  \item RNN \cite{graves2012sequence}: a recurrent neural network designed to iteratively process sequential data.
  \item CNN \cite{kim2014convolutional}: a 1D convolutional neural network for sequence modeling.
  \item TCN \cite{zhang2022production}: a temporal dilated convolutional network for sequence modeling.
  \item GCN \cite{kipf2016semi}: a graph convolutional neural network for time series forecasting.
  \item Transformer \cite{vaswani2017attention}: an attention-based network for sequence modeling. 
\end{itemize}

\paragraph{Prediction errors.} 
The mean average errors (MAEs) and root mean square errors (RMSEs) were the two measures used to evaluate these models.

\section{Results and Discussion}
Table \ref{tab:predictionerrors} shows MAE and RMSE scores on the test set for each method used. Each experiment was performed with and without the future covariates, mirroring the situation when reliable predictions are or are not available for the computation.

The results indicate that the use of reliably predicted future information can improve the accuracy of predictions.
In addition, our FloodGTN-Parallel outperformed other baseline models in this paper while FloodGTN-Series had a performance level comparable to the best baseline.

\begin{table}[ht]
\caption{Average 24-hour prediction errors on test set with/without future predicted covariates (F.P.C).}
\label{tab:predictionerrors}
\centering
\begin{tabular}{lllll}
\toprule
\multicolumn{1}{c}{} & \multicolumn{2}{l}{With F.P.C.}   & \multicolumn{2}{l}{Without F.P.C.}       \\
\cmidrule(r){2-3}     \cmidrule(r){4-5}
Model               & MAE (ft)         & RMSE (ft)       & MAE (ft)         & RMSE (ft)  \\
\midrule\midrule
HEC-RAS             & 0.185            & 0.238            & 0.185           & 0.238  \\
\midrule
RNN                 & 0.087            & 0.112            & 0.167           & 0.213  \\
CNN                 & 0.105            & 0.135            & 0.203           & 0.265  \\
TCN                 & 0.085            & 0.114            & 0.162           & 0.207  \\
GCN                 & 0.075            & 0.102            & 0.138           & 0.183  \\
Transformer         & 0.068            & 0.091            & 0.132           & 0.175 \\
\midrule
FloodGTN-Series    & 0.069            & 0.094            & 0.135           & 0.179  \\
FloodGTN-Parallel  & \textbf{0.055}   & \textbf{0.079}   & \textbf{0.127}  & \textbf{0.169} \\
\bottomrule
\end{tabular}
\end{table}

\begin{figure}[ht]
\centering
\includegraphics[scale=0.38]{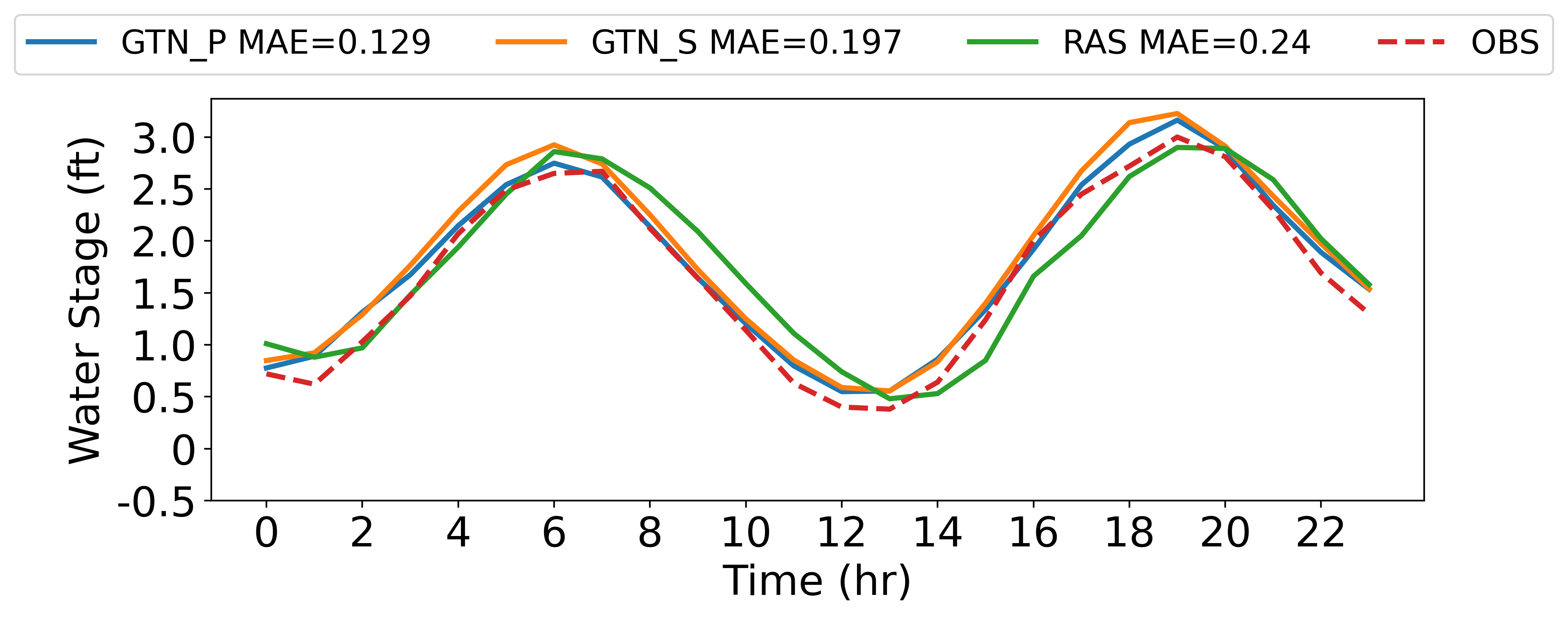}
\caption{Visualization of observed \& predicted water levels with future predicted covariates. GTN\_P, GTN\_S, and RAS represent FloodGTN-Parallel, FloodGTN-Series, and HEC-RAS models, respectively. OBS denotes the observed data.}
\label{fig:visualize}
\end{figure}

\paragraph{Visualization.} Figure \ref{fig:visualize} shows the observed and predicted water levels of some selected models during the tropical storm \emph{Isalas} that happened on August 1st, 2020.

\paragraph{Computational time.} To compare the efficiency of the physics-based models and DL models, we show the simulation time of HEC-RAS and the prediction time of DL models on the test set. Note that the DL models are pre-trained beforehand and only applied during the test phase.

\begin{table}[ht]
\centering
  \caption{Simulation time of physics-based HEC-RAS and the prediction time of DL models.}
  \label{tab:predictiontime}
  \begin{tabular}{lllllllll}
    \toprule
    Model            & HEC-RAS     & RNN      & CNN        & TCN       & GCN       & Transf.   & GTN\_S      & GTN\_P  \\
    \midrule\midrule
    Train       & -           & 427 min    & 29 min     & 366 min   & 60 min    & 47 min    & 686 min    & 148 min    \\
    Test        & 45.92 min   & 4.81 s    & 1.81 s     & 4.95 s    & 4.04 s    & 4.46 s    & 5.51 s    & 5.02 s   \\
    \bottomrule
  \end{tabular}
\end{table}


\section{Conclusions}
We propose a Graph Transformer Network for Flood prediction called \texttt{FloodGTN}. 
From the accuracy perspective, both \texttt{FloodGTN-Series} and \texttt{FloodGTN-Parallel} models outperform the physics-based \texttt{HEC-RAS} model, and \texttt{FloodGTN-Parallel} is better than all other DL-based methods.
More importantly, once trained, the prediction speed of the DL models is much faster than the simulation speed of physics-based \texttt{HEC-RAS}.
This opens the door to the possibility of a high-quality real-time prediction tool, the first step toward building a sophisticated flood management system.
We also demonstrate that using future exogenous covariates (such as rainfall and tide) is critical for DL models to produce more accurate predictions.

\begin{ack}
This work is part of the I-GUIDE project, which is funded by the National Science Foundation under award number 2118329.
\end{ack}


\bibliographystyle{plain}
\bibliography{references}


\end{document}